\documentclass[11pt]{article}
\usepackage{etex}
\usepackage[preprint]{acl}

\usepackage{times}
\usepackage{latexsym}

\usepackage[T1]{fontenc}

\usepackage[utf8]{inputenc}

\usepackage{microtype}

\usepackage{inconsolata}

\usepackage{graphicx}

\usepackage{booktabs}
\usepackage{listings}

\usepackage{xcolor}
\lstdefinelanguage{yaml}{
  morestring=[b]',
  morestring=[b]",
  sensitive=false,
  morecomment=[l]{\#},
  commentstyle=\color{gray}\ttfamily,
  stringstyle=\color{teal}\ttfamily,
  morekeywords={true,false,null,yes,no,on,off},
  literate=
   *{:}{{:}}{1}
    {-}{{-}}{1}
    {>}{{>}}{1}
    {|}{{|}}{1}
}

\usepackage{amsmath}
\usepackage[english]{babel}
\usepackage{hyperref}
\addto\captionsenglish{%
}
\addto\extrasenglish{%
}

\addto\captionsenglish{%
}
\usepackage{multirow} 

\newcommand{\GuidelineBlock}[4]{%
  \IfFileExists{#1}{%
    \subsection{#2}\label{subsec:#3}
    \noindent\textit{#4}

    \begin{minipage}[t]{0.98\textwidth}
    \vspace{4pt}
    \lstinputlisting[
      caption={#2},label={lst:#3}
    ]{#1}
    \end{minipage}
    \vspace{1.0em}
  }{}
}
\usepackage{enumitem} 

\title{Evaluating the Impact of Reviewer Guideline Design \\on LLM-Based Automated Peer Review}

\author{
  Haowen Li\textsuperscript{1}\thanks{Work done during an internship at NEC Corporation.}, 
  Yoichi Ishibashi\textsuperscript{2}, 
  Masafumi Oyamada\textsuperscript{2} \\
  \textsuperscript{1}Keio University, 
  \textsuperscript{2}NEC Corporation \\
  \texttt{tony\_li.haowen@keio.jp},
  \texttt{\{yoichi-ishibashi, oyamada\}@nec.com}
}

\begin{document}
\maketitle

\begin{abstract}
Peer review is an essential process in scientific research, yet the growing workload has made its automation increasingly necessary. In this study, we analyze how different types of \emph{reviewer guidelines}, such as official conference guidelines and reviewer-imitating ones generated from high-quality human reviews using LLMs, affect automated peer review. Our experiments show that official conference guidelines produce review results most consistent with human judgments, suggesting that evaluation criteria refined through conference practice serve as effective guidance for automated reviewing as well. In contrast, reviewer-imitating guidelines were generally less effective than official conference guidelines. Furthermore, enforcing strict rubric-style scoring consistently degraded performance, highlighting the importance of allowing subjective and holistic scoring.
\end{abstract}

\section{Introduction}
In recent years, the explosive growth in submissions to major machine learning conferences has placed an enormous burden on peer reviewers~\cite{stelmakh2020novicereviewerexperimentaddressscarcity, zhang2022, wei2025aiimperativescalinghighquality, kim2025positionaiconferencepeer}. The current human-based peer review system faces serious challenges including reviewer overload, inconsistent judgments, superficial evaluations, and delayed feedback. At the same time, as research automation advances, the importance of accurate evaluation has only increased, making the automation of the peer review process an urgent necessity.  

One clear illustration comes from the ACL Rolling Review (ARR) guidelines, which ask reviewers to ``\emph{Check for common review issues}''\footnote{\url{https://aclrollingreview.org/reviewerguidelines}}. These items do not simply caution reviewers against mistakes, but establish concrete criteria for judging the scientific contribution of a paper: for example, a result should not be undervalued merely because it appears ``unsurprising,'' nor should a contribution be dismissed solely because it contradicts prior beliefs. By formalizing such principles, ARR effectively provides operational indicators that define what counts as sound, novel, and impactful research. Such directions illustrate how, as noted by \citet{seeber2020journals}, reviewer guidelines shape the nature and quality of peer review rather than serving as administrative notes.

While several studies have explored automated peer review using Large language models (LLMs)~\cite{zhuang2025largelanguagemodelsautomated, shin2025mindblindspotsfocuslevel, chen2025, lin2025breakingreviewerassessingvulnerability, li2025unveilingmeritsdefectsllms, weng2025cycleresearcherimprovingautomatedresearch}, the impact of guidelines and rubrics that govern review policies on automated review performance has not been sufficiently investigated. Guidelines direct reviewers’ attention to particular aspects and perspectives when evaluating a paper, and thus are crucial determinants of review quality. However, it remains unclear what types of guidelines are effective, and what types are ineffective, for automated peer review.  

Broadly, two approaches can be considered for guideline design in automated peer review. One is to directly provide LLMs with existing, human-written guidelines established for each conference (e.g., official reviewer guideline from ARR) and have them conduct reviews. The other is a reviewer-imitating approach, where high-quality human-written reviews are used as input to a guideline generator (such as an LLM), which then extracts recurring evaluation patterns and key perspectives to generate new guidelines.  

In this study, we systematically investigate how review quality is affected when LLMs conduct reviews under different guideline conditions. Specifically, we address the following research questions:

\noindent\textbf{RQ1: Does providing guidelines improve the performance of automated peer review?} \\
\textbf{A:} We found that providing guidelines improves the performance of automated peer review. In particular, using official conference guidelines, such as those from NeurIPS and ARR, proved to be more effective. This suggests that evaluation criteria refined over time by the research community serve as a strong foundation for automated reviewing as well (\autoref{sec:conf_guidelines}).

\vspace{0.2cm}

\noindent\textbf{RQ2: Are reviewer-imitating guidelines effective? In particular, is it better for models to determine scores subjectively, or to compute them strictly using rubric-style criteria?} \\
\textbf{A:} We found that reviewer-imitating guidelines were generally less effective than official conference guidelines. In particular, rubric-style formulations further reduced alignment with human judgments. This suggests that allowing LLMs to determine scores subjectively may offer better alignment with human judgment in specific contexts, rather than enforcing rigid, additive scoring logic~(\autoref{sec:reviewer_imitating}).

\section{Problem Formulation}
\paragraph{Types of Reviewer Guidelines}
This study analyzes two approaches for automated peer review:  
(1) using human-created guidelines (e.g., official ICLR reviewer guidelines) directly provided to LLMs for reviewing, and (2) generating guidelines from high-quality human reviews (filtered based on criteria in \autoref{sec:guideline-description}) using LLMs and then using these generated guidelines for review.

\paragraph{Guideline-Based Automated Review}
Both approaches follow a common input–output structure for automated review, as shown in the prompt template in Appendix (\autoref{lst:peer-review-prompt}).
The \textbf{input} consists of: (i) a paper in text format\footnote{Converted from PDF with references and appendices removed}, and (ii) review guidelines (either human-created or LLM-generated). The \textbf{output} is a numerical review score (typically on a 1-10 scale) along with textual justification.

\paragraph{Evaluation}
To evaluate the quality of automated reviews, we compute the root mean squared error (RMSE) between LLM-generated scores and human review scores for the same papers. A lower RMSE indicates that the model’s ratings are numerically closer to human ratings. Given our 1-10 scoring scale, an RMSE of 3.0 indicates that the model's prediction deviates from the human average by approximately 3 points. This metric allows us to assess how effectively an automated system approximates human peer review~\cite{wei2025aiimperativescalinghighquality}. While RMSE serves as our primary quantitative metric, a qualitative textual analysis of the generated justifications is provided in Appendix D to examine the logical depth and actionability of the reviews.

\section{Effectiveness of Conference Reviewer Guidelines for Automated Review}
\label{sec:conf_guidelines}

This section evaluates how well reviewer guidelines improve automated peer review. We address whether providing guidelines improves accuracy and which venue’s guidelines are most effective.

\begin{table}[t]
\small
\centering
\setlength{\tabcolsep}{0.8mm}
\begin{tabular}{lcccc}
\toprule
Model & No-Guideline & ARR & NeurIPS & ICLR \\
\midrule
\multicolumn{5}{c}{\emph{Official conference guidelines}}\\
\midrule 
Qwen3-30B-A3B        & 3.06 & 3.18 & \textbf{2.07} & \underline{2.91} \\
DeepSeek-R1-32B     & 3.86 & 3.27 & \textbf{2.72} & \underline{2.91} \\
Qwen3-32B            & 3.58 & 3.32 & \textbf{2.51} & \underline{2.89} \\
\addlinespace[2pt]
\midrule
\multicolumn{5}{c}{\emph{Guidelines rewritten into instruction-style prompts}}\\
\midrule
Qwen3-30B-A3B        & 3.06 & \textbf{2.97} & 3.08 & \underline{2.99} \\
DeepSeek-R1-32B     & 3.86 & \underline{3.06} & 3.18 & \textbf{3.05} \\
Qwen3-32B            & 3.58 & \underline{3.16} & 3.32 & \textbf{3.09} \\
\bottomrule
\end{tabular}
\caption{RMSE of automated review scores under No-Guideline and official conference guidelines. Upper block shows results using  \emph{official reviewer guidelines}, while lower block uses the same content rewritten into \emph{instruction-style prompts} (converted with ChatGPT). Best results are in \textbf{bold}, second-best are \underline{underlined}. \textbf{Lower values are better}, indicating closer alignment with human scores. }
\label{tab:conference_guidelines}
\end{table}

\paragraph{Setup}
We randomly sampled 1{,}500 submissions (papers and reviews) from the ICLR~2024 peer-review dataset\footnote{We collected data from ICLR 2024 submissions on OpenReview, including 7,262 paper PDFs and 28,028 open-access reviews from \url{https://openreview.net/group?id=ICLR.cc/2024/Conference}}. For each paper, the human reference label is the average of its reviewers' overall rating. We evaluate three models: Qwen3-30B-A3B (MoE model)~\cite{DBLP:journals/corr/abs-2505-09388}, DeepSeek-R1-Distill-Qwen3-32B~\cite{deepseekai2025deepseekr1incentivizingreasoningcapability}, and Qwen3-32B~\cite{DBLP:journals/corr/abs-2505-09388}. We intentionally prioritize these open-weights models over proprietary state-of-the-art systems (e.g., GPT-5 or Gemini 3) to ensure a scientifically transparent, reproducible, and cost-effective foundation for the research community. While evaluating the absolute performance ceiling with proprietary models is valuable, processing 1{,}500 full papers requires massive token consumption, and the "black box" nature of commercial APIs makes their shifting behaviors difficult to reproduce. Because our primary objective is to establish the relative impact of different guideline designs, utilizing capable open-weights models effectively isolates the effect of the guidelines without compromising the integrity of our core findings.
``No-Guideline'' denotes prompting the model with the paper only (no guideline) to output an integer overall rating, as shown in \autoref{sec:no-guideline}. This serves as a strict, zero-shot lower bound; while the prompt assigns an ``expert'' persona, it intentionally lacks any specific evaluative context or structural requirements to isolate the pure impact of the guidelines. We also evaluate using official reviewer guidelines from ICLR\footnote{\url{https://iclr.cc/Conferences/2024/ReviewerGuide}}/NeurIPS\footnote{\url{https://neurips.cc/Conferences/2024/ReviewerGuidelines}}/ARR\footnote{\url{https://aclrollingreview.org/reviewerguidelines}}. Because official documents often include non-evaluative administrative notes and explanatory prose, we additionally test an instruction-style variant where the same evaluative content is rewritten as imperative, bullet-pointed instructions with explicit output requirements converted using ChatGPT 5 (GPT-5). 
The prompts used for all guideline-based conditions is shown in \autoref{sec:prompt}.

\paragraph{Q1: Do guidelines improve automated reviewing?}
We compare No-Guideline to runs with conference guidelines. As shown in \autoref{tab:conference_guidelines}, RMSE consistently decreases across all models when guidelines are provided. For example, DeepSeek-R1-32B demonstrates a substantial reduction in error across all three conference venues compared to the No-guideline baseline. Similarly, the instruction-style prompts outperform the baseline for all tested models, indicating that explicit evaluative criteria improve agreement with human judgements.

\paragraph{Q2: Which conference guidelines are more effective?}
We next compare ICLR/NeurIPS/ARR against each other. With the official guideline text, all three models achieve the lowest RMSE under NeurIPS, followed by ICLR, then ARR (e.g., for Qwen3-30B-A3B: 2.07 $<$ 2.91 $<$ 3.18). When the same content is converted into instruction-style prompts, the best venue varies by model: Qwen3-30B-A3B prefers ARR (2.97), while DeepSeek-R1-32B and Qwen3-32B prefer ICLR (3.05 and 3.09, respectively). In other words, beyond the guideline content, the presentation (verbatim policy text vs.\ instruction-style prompt) can alter model behavior and flip the relative ranking among venues.

\section{Effectiveness of Reviewer-imitating Guidelines in Automated Peer Review}
\label{sec:reviewer_imitating}

In addition to using human-written conference reviewer guidelines, another approach is to \emph{construct} guidelines with an LLM. In this section, we examine whether \emph{reviewer-imitating guidelines}, generated from human-written reviews, can improve automated peer review. We evaluate review quality using several indicators (such as review length) and extract common perspectives from high-quality reviews, which are then converted into guidelines by an LLM. Through this filtering process, we divide reviews into three groups: \emph{Good}, \emph{Middle}, and \emph{Bad Review}. We generate guidelines from each group and compare automated review performance under these different settings.

\paragraph{Overview of Guideline Generation}
We generate reviewer-imitating guidelines and apply them to automated peer review through the following steps:

\begin{lstlisting}[language=yaml, caption={Example of reviewer-imitating guideline extracted from high-quality (``Good'') reviews.}, label={lst:yaml-guideline}]
Provide the following based on the Peer Review Behavior Checklist:

peer_review_behavior_checklist:
    good_paper:
    contribution_and_impact:
      - question: "Does the paper tackle a significant and timely problem?"
        description: "Reviewers positively note papers that address important and relevant research topics, seeing it as a sign of potential impact."
      - question: ...
    clarity_and_presentation: ...
    bad_paper:
    novelty_and_scholarship:
      - question: "Is the contribution incremental or a re-implementation of existing ideas?"
        description: "This is a frequent and critical flaw. Reviewers are highly knowledgeable and will reject papers they deem to be minor variations of prior work, often citing the specific papers."
      - question: ...
    clarity_and_soundness: ...
\end{lstlisting}

\begin{itemize}
    \item \textbf{STEP1: Review Data Filtering}:  
    To ensure guidelines are distilled from thorough reviews, we filter data using programmatic proxies: Length, Submission Date, Citations, and Consistency. While manual expert annotation is a gold standard, these structural metrics provide a scalable and reproducible foundation for assessing reviewer engagement and thoroughness. Reviews are grouped into \emph{Good}, \emph{Middle}, and \emph{Bad} based on a composite score ($3$-$17$ points).
    \item \textbf{STEP2: Guideline Generation}:  
    We provide the filtered reviews to Gemini 2.5 Pro \cite{Gemini2_5_2025} to generate reviewer-imitating guidelines. Specifically, we randomly sampled exactly 30 reviews from each of the three quality pools (Good, Middle, and Bad) defined in Step 1 to generate corresponding reviewer-imitating guidelines. The resulting guidelines are structured as a checklist, with separate items for positive aspects (good papers) and negative aspects (bad papers). They capture frequently mentioned criticisms, recurring evaluation criteria, and characteristic perspectives. For example, \autoref{lst:yaml-guideline} shows an actual generated guideline (excerpted for brevity). Further details are described in \autoref{sec:guideline_generation}.
    \item \textbf{STEP3: Automated Review Using Guidelines}: 
    This step follows the same procedure as in the conference guideline experiments. The generated reviewer-imitating guidelines are given to the LLMs together with the paper text. The model then outputs an overall rating on a 1–10 scale and gives a brief justification.
\end{itemize}

\begin{table}[t]
\small
\centering
\setlength{\tabcolsep}{1.1mm} 
\resizebox{\columnwidth}{!}{ 
\begin{tabular}{l l c c c c}
\toprule
Model & Setting      & Base. & Good & Mid. & Bad \\
\midrule
\multirow{2}{*}{Qwen3-30B-A3B}
 & w/o Rubric & \textbf{3.06} & 3.22 & 3.24 & \underline{3.14} \\
 & w/  Rubric & N/A  &  4.42   & \textbf{4.10} & \underline{4.28} \\
\midrule
\multirow{2}{*}{DeepSeek-R1-32B}
 & w/o Rubric & 3.86  & 3.13               & \underline{3.12} & \textbf{3.06} \\
 & w/  Rubric & N/A  & \underline{4.26}  & \textbf{3.92}    & \underline{4.26} \\
\midrule
\multirow{2}{*}{Qwen3-32B}
 & w/o Rubric & 3.58 & \textbf{3.48}     & \underline{3.51} & 3.69 \\
 & w/  Rubric & N/A  & \underline{4.57}  & \textbf{4.31}    & 4.62 \\
\bottomrule
\end{tabular}
}
\caption{RMSE (lower values are better) under reviewer-imitating guidelines, comparing settings with and without rubric. ``Good'' and ``Bad'' denote guidelines distilled from \emph{high-quality} and \emph{low-quality} human reviews, respectively.}
\label{tab:reviewer_imitating}
\end{table}

\paragraph{Q1: Do reviewer-imitating guidelines outperform No-Guideline?}
We compare No-Guideline (the model uses only the paper) against reviewer-imitating guidelines from the Good/Bad groups. The impact of reviewer-imitating guidelines is model-dependent (see \autoref{tab:reviewer_imitating} for the comparison with the No-Guideline baseline). While DeepSeek-R1-32B shows a notable improvement in alignment, other models such as Qwen3-30B-A3B experience performance degradation when using guidelines distilled from human reviews. This suggests that reviewer imitation does not yield the same uniform gains observed with official conference guidelines.

\paragraph{Q2: How do reviewer-imitating guidelines compare to conference guidelines?}
We compare reviewer-imitating guidelines (Good/Bad) to official conference reviewer guidelines (ICLR, NeurIPS, ARR) under the same evaluation setup. Cross-referencing \autoref{tab:conference_guidelines}, conference guidelines consistently achieve lower RMSE than reviewer imitation. For example, with Qwen3-30B-A3B, Good = 3.22 and Bad = 3.14, while NeurIPS = 2.07 and ICLR = 2.91. These results suggest that the structural clarity and explicit requirements in official guidelines calibrate models more effectively than behavior-only signals distilled from reviews. Qualitatively, the guidelines themselves differ significantly: official guidelines mandate structural completeness (e.g., dedicated ethics and limitations sections), whereas reviewer-imitating guidelines heavily index on behavioral tropes such as 'missing ablations' or 'incremental novelty' (see \autoref{sec:guideline_generation}).

\paragraph{Q3: Does rubricizing reviewer-imitating guidelines help?}
We compare free-form (non-rubric) instructions to rubric-style scoring for reviewer-imitating guidelines. 
In the rubric setting, the LLM judges whether each checklist item (\autoref{lst:yaml-guideline}) in the ``good\_paper'' section is satisfied and assigns +1 point if so, while items in the ``bad\_paper'' section contribute -1 point. The total score is then normalized to a 1–10 scale and used as the overall rating.
As shown in \autoref{tab:reviewer_imitating}, the non-rubric setting consistently outperforms rubric-style formulation across all models and quality groups. In all cases, enforcing rigid point-allocation system significantly increased RMSE compared to allowing the model to provide free-form justifications.

\section{Related Work}
Automatic peer review research can be broadly categorized into two approaches: learning-based and prompt-based.  
The learning-based approach trains models to generate reviews or predict scores using task-specific supervision and intermediate reasoning.  
Early studies generated explainable reviews from structured evidence built through knowledge graphs~\cite{gao-etal-2020-reviewrobot}.  
More recent work employs multi-stage pipelines that emulate expert analytical processes to train end-to-end review models~\cite{zhu2025deepreview}.  
In addition, recent surveys have organized datasets and tasks, helping to systematize what is learned and how models are evaluated~\cite{yuan2022automate,dycke-etal-2023-nlpeer}.  
The prompt-based approach guides large language models during inference using rubrics or instructions.  
Studies on prompt-based automated reviewing have shown that well-designed evaluation criteria improve the fidelity and granularity of judgments~\cite{kim2023prometheus}.  
Benchmarking frameworks further standardize prompt formats and evaluation procedures~\cite{zheng2023mtbench}.  
These studies collectively highlight the importance of explicit instructions for LLMs.  
Building on this insight, we analyze how using reviewer guidelines, which serve as instructions to human reviewers, as prompts affects the quality of automated peer review.  
A related study~\cite{DBLP:journals/corr/abs-2502-11736} uses existing conference guidelines to generate and evaluate LLM-generated reviews.
In contrast, we not only compare official guidelines from multiple venues but also propose generating reviewer-imitating guidelines from human reviews using LLMs, and systematically evaluate how these different guideline types affect scoring alignment with human judgments

\label{sec:bibtex}

\section*{Limitations}
This study is limited in several aspects. First, we only analyze review data written in English, as the ICLR 2024 dataset contains exclusively English reviews. Consequently, our findings may not generalize to non-English scientific contexts where linguistic structures and cultural factors influence evaluation criteria.

Furthermore, our evaluation relies primarily on RMSE between LLM-generated scores and human averages. While we supplement this with a qualitative textual analysis (Appendix~\ref{sec:analysis}) and score distribution visualizations (Appendix~\ref{sec:score_distributions}), our textual analysis is limited to a single case study and does not include systematic quantitative measures such as vocabulary richness or type-token ratio. Extending these analyses across a larger sample of papers remains an important direction for future work.

Finally, while our reviewer-imitating guidelines are derived from high-quality human reviews, their effectiveness depends on the specific logic and potential biases of the generator model (Gemini 2.5 Pro) and the structural metrics used for filtering. Future research should explore cross-model guideline generation and human-in-the-loop validation to ensure that automated criteria remain aligned with evolving scientific standards.

\section*{Ethics Statement}
As automated peer review systems increasingly rely on Large Language Models (LLMs), a critical ethical concern is the preservation of manuscript confidentiality. The current peer review system relies on strict non-disclosure agreements to protect the intellectual property of authors before publication. Therefore, researchers, reviewers, and meta-reviewers must exercise extreme caution to avoid uploading unpublished, confidential submissions to non-privacy-preserving commercial APIs (such as the standard web interfaces of ChatGPT, Gemini, or Claude). Inputting unpublished manuscripts into these platforms risks violating conference confidentiality policies, as the data may be retained or used in future model training pipelines. The development and deployment of LLM-based review assistants must strictly mandate the use of privacy-preserving tools, local models, secure enterprise API endpoints, or platforms with explicit zero-data-retention agreements to ensure the integrity, privacy, and trust of the peer review process.

\bibliography{custom}

\appendix

\section{Detailed Description of the Reviewer-imitating Guideline}
\label{sec:guideline-description}
\subsection{Review Data Collection and Filtering}
\label{sec:dataset}

We collected data from ICLR 2024 submissions on OpenReview, including 7,262 paper PDFs and 28,028 open-access reviews as mentioned in \autoref{sec:conf_guidelines}. We selected ICLR due to the abundance of open-access data and its status as a representative conference in the machine learning field.
\autoref{tab:review_example} shows an example of review data used in our study. Each review includes the following elements: review summary, item-specific scores (soundness, presentation, contribution), strengths, weaknesses, questions, overall score, and confidence level. 
However, not all reviews are of high quality. Some reviews are too short, have unclear evidence, or have inconsistent scores and content. To generate guidelines from high-quality reviews, it is necessary to filter the review data as preprocessing.
Filtering uses the following five criteria, based on our intuitions about what constitutes high-quality reviews.

\begin{table*}[h]
\small
\centering
\begin{tabular}{p{3cm}p{9cm}}
\toprule
\textbf{Field} & \textbf{Content} \\
\midrule
Summary & \textit{This paper proposes a method for multimodal learning...} \\
\midrule
Strengths & \textit{(1) The motivation is clear... (2) ...} \\
\midrule
Weaknesses & \textit{(1) The novelty is limited... (2) ...} \\
\midrule
Questions & \textit{(1) How does the method scale with model size?... (2) ...} \\
\midrule
Soundness & 3 (good) \\
\midrule
Presentation & 3 (good) \\
\midrule
Contribution & 3 (good) \\
\midrule
Overall Rating & 8: accept, good paper \\
\midrule
Confidence & 4: You are confident in your assessment, .. \\
\midrule
Submission Date & \{Unix timestamp\} \\
\bottomrule
\end{tabular}
\caption{Illustrative example of review data inspired by the ICLR~2024 peer review dataset (not from actual reviews). 
The content is paraphrased and does not reproduce any actual review text. 
To extract high-quality reviews for guideline generation, we filter this data using five predefined criteria: 
\textbf{(1) Submission Date} (uses timestamp to measure timing appropriateness), 
\textbf{(2) Review Length} (evaluates character count from textual fields), 
\textbf{(3) Reference Quality} (analyzes citation mentions in textual content), 
\textbf{(4) Content Consistency} (assesses alignment between textual tone and numerical scores), 
and \textbf{(5) Subscore Consistency} (checks coherence between item-specific and overall ratings). 
Our experiments also analyze which filtering criteria are most effective for extracting quality reviews.}
\label{tab:review_example}
\end{table*}

\begin{enumerate}
    \item \textbf{Submission Date}: Measures deviation from the average review submission date. We hypothesize that reviewers who submit reviews promptly are more conscientious and likely to produce higher quality reviews.
    \item \textbf{Review Length}: Evaluation based on character count, with longer reviews assumed to indicate deeper engagement with the paper. We calculate scores based on multiples of the median length.
    \item \textbf{Reference Quality}: Evaluates mention of one or more references, pointing out insufficient related work, and critical discussion of prior work (0-3 points). We expect that reviewers with higher domain expertise, evidenced by reference to relevant literature, produce better reviews. We use LLMs to score this criterion by inputting review data and prompting for assessment of reference usage.
    \item \textbf{Content Consistency}: Evaluates consistency between review text tone and evaluation score, and alignment between justification and score (0-2 points). We assume that reviews with inconsistent tone and ratings (e.g., positive text with low scores or vice versa) indicate poor review quality. We employ LLMs to assess this consistency by analyzing the alignment between textual content and numerical scores.
    \item \textbf{Subscore Consistency}: Evaluates consistency between item scores (soundness, presentation, contribution) and overall score, and alignment between each subscore and evaluation content (0-2 points). Similar to content consistency, we expect coherent scoring patterns to indicate higher review quality. LLMs evaluate this criterion by examining the coherence between different scoring components.
\end{enumerate}

Filtering scores are distributed in the range of 3-17 points, with a median of 11 points. Based on this score, we create high-quality (``Good''), medium-quality (``Middle''), and low-quality (``Bad'') review groups.
We assume that reviews satisfying these criteria represent high-quality reviews. However, we empirically analyze whether these assumptions hold true and investigate which criteria positively (or potentially negatively, contrary to our intuition) correlate with automated review performance in our experiments.


\subsection{Filtering Criteria Details}

This section provides detailed calculation procedures for the five filtering criteria used to assess review quality.

\paragraph{Submission Date}

The submission date criterion evaluates the timing of review submission relative to the conference's review timeline. We calculate the deviation from the average review submission time across all reviews for the same conference.
Let $t_i$ be the submission timestamp for review $i$, and $\bar{t}$ be the average submission time across all reviews. The deviation score is calculated as $d_i = |t_i - \bar{t}|$.
Reviews are scored based on their deviation from the average submission time:
\begin{itemize}
    \item +5 point: $d_i \leq 2$ days from average
    \item +4 point: $2 < d_i \leq 5$ days from average
    \item +3 point: $5 < d_i \leq 10$ days from average
    \item +2 point: $10 < d_i \leq 20$ days from average
    \item +1 point: $d_i > 20$ days from average
\end{itemize}

\paragraph{Review Length}
The review length criterion evaluates the total character count of the review text, including summary, strengths, weaknesses, and questions sections.
Let $l_i$ be the character count for review $i$, and $M$ be the median character count across all reviews. The length score is determined as follows:
\begin{itemize}
    \item +5 point: $l_i > 1.5 \times M$
    \item +4 point: $1.2 \times M < l_i \leq 1.5 \times M$
    \item +3 point: $0.8 \times M < l_i \leq 1.2 \times M$
    \item +2 point: $0.5 \times M < l_i \leq 0.8 \times M$
    \item +1 point: $l_i \leq 0.5 \times M$
\end{itemize}

\paragraph{Reference Quality}
Reference quality is evaluated using LLM assessment of how well the review engages with relevant literature. The scoring criteria (0-3 points) are:
\begin{itemize}
    \item +1 point: Review mentions one or more relevant references
    \item +1 point: Review identifies missing or insufficient related work
    \item +1 point: Review provides critical discussion of prior work
\end{itemize}
The full prompt is shown in \autoref{lst:reference-quality}.
\begin{lstlisting}[caption={Prompts of reference quality assessment}, label={lst:reference-quality}]

Your task is to evaluate the quality of a peer review. Specifically, analyze whether the reviewer suggests any additional references or points out missing ones.

# REVIEW CONTENT:
Summary: {summary}
Strengths: {strengths}
Weaknesses: {weaknesses}
Questions: {questions}

# TASK:

# Step 1: Analysis  
Evaluate the review using the rubric below. For each applicable item, add **+1 point** (maximum: 3 points).

## Rubric:
- The reviewer explicitly suggests or mentions one or more references (+1 point)  
- The reviewer identifies missing citations or insufficient coverage of related work (+1 point)  
- The reviewer discusses prior work in a detailed or critical way (+1 point)  

## Briefly explain which of the above items apply and why.

# Step 2: Output  
Based on the total score, assign `reference_quality` as follows:  
- 3 or 2 points : `"high"`  
- 1 point : `"medium"`  
- 0 points : `"low"`  

Then summarize your judgment in the following **YAML** format:
```yaml
"suggests_references": true/false,
"reference_count": integer,
"points_out_missing_citations": true/false,
"quality_score": 0/1/2/3,
"reasoning": "brief explanation of what references were found or what gaps were pointed out",
"reference_quality": "high" / "medium" / "low"
```
\end{lstlisting}

\paragraph{Content Consistency}
Content consistency measures the alignment between the textual content of the review and the numerical scores provided. This is assessed using LLMs with the following criteria (0-2 points):
\begin{itemize}
    \item +1 point: Review tone matches the numerical rating (positive tone with high scores, critical tone with low scores)
    \item +1 point: Justification provided in the text aligns with the overall score
\end{itemize}
The full prompt is shown in \autoref{lst:content-consistency}.
\begin{lstlisting}[caption={Prompts of content consistency assessment}, label={lst:content-consistency}]

Your task is to evaluate the consistency between a reviewer's **overall rating** and the **textual content** of their review (e.g., strengths, weaknesses, and summary).

# REVIEW:
Rating: {rating}

## Summary:
{summary}

## Strengths:
{strengths}

## Weaknesses:
{weaknesses}

# TASK:

## Step 1: Analysis
Evaluate the review using the rubric below. For each applicable item, add **+1 point** (maximum: 2 points).

### Rubric:
- The **overall tone** of the review (as reflected in strengths/weaknesses) matches the numerical rating (+1 point)  
  _(e.g., a review with mostly strong praise and minor weaknesses aligns with a high rating like "8: accept")_

- The **justifications provided in text** reasonably support the rating given, showing a clear rationale (+1 point)  
  _(e.g., the rating is backed up with concrete positive or negative points related to novelty, results, impact, etc.)_

## Step 2: Output
Based on the total score, assign `content_consistency` as follows:
- 2 points: "consistent"
- 1 point: "somewhat inconsistent"
- 0 points: "inconsistent"

Then summarize your judgment in the following **YAML** format:

```yaml
rating: "{rating}"
summary_excerpt: short excerpt (1-2 sentences) summarizing the review
key_strengths: brief key points reflecting strengths
key_weaknesses: brief key points reflecting weaknesses
quality_score: 0/1/2
reasoning: "brief explanation of how the textual content aligns (or not) with the rating"
content_consistency: "consistent" / "somewhat inconsistent" / "inconsistent"
\end{lstlisting}

\paragraph{Subscore Consistency}
Subscore consistency evaluates the coherence between item-specific scores (soundness, presentation, contribution) and the overall rating. LLMs assess this using the following criteria (0-2 points):
\begin{itemize}
    \item +1 point: Overall score reasonably reflects the average of subscores
    \item +1 point: Each subscore aligns with the corresponding evaluation aspects discussed in the text
\end{itemize}
The full prompt is shown in \autoref{lst:subscore-consistency}.

\begin{lstlisting}[caption={Prompts of subscore consistency assessment}, label={lst:subscore-consistency}]

Your task is to evaluate the consistency between a reviewer's overall rating and the subscores they assigned.

# REVIEW SCORES:
Rating: {rating}
Soundness: {soundness}
Presentation: {presentation}
Contribution: {contribution}

# TASK:

# Step 1: Analysis
Evaluate the review using the rubric below. For each applicable item, add **+1 point** (maximum: 2 points).

## Rubric:
- The overall rating is in line with the average or weighted impact of the subscores (+1 point)
- The subscores reflect distinct and plausible dimensions of evaluation (e.g., a paper with strong contribution but weak presentation is rated moderately overall) (+1 point)

## Briefly explain which of the above items apply and why.

# Step 2: Output
Based on the total score, assign subscore_consistency as follows:
- 2 points: "consistent"
- 1 point: "somewhat inconsistent"
- 0 points: "inconsistent"

Then summarize your judgment in the following **YAML** format:
```yaml
"rating": "{rating}",
"soundness_score": "{soundness}",
"presentation_score": "{presentation}",
"contribution_score": "{contribution}",
"average_subscore": float,
"quality_score": 0/1/2,
"reasoning": "brief explanation of whether the subscores align or conflict with the overall rating",
"subscore_consistency": "consistent" / "somewhat inconsistent" / "inconsistent"
```
\end{lstlisting}

\paragraph{Score Distribution}
The combined filtering scores range from 3 to 17 points (sum of all five criteria). 
To analyze review quality, we sampled three groups of reviews based on their scores:
\begin{itemize}
    \item \textbf{Good}: 30 reviews randomly sampled from those with a score of 17
    \item \textbf{Middle}: 30 reviews randomly sampled from those with a medium score of 11
    \item \textbf{Bad}: 30 reviews randomly sampled from those with scores of 6 or lower
\end{itemize}

\subsection{Guideline Generation}
\label{sec:guideline_generation}

We concatenate the filtered reviews into a single text prompt and input it to Gemini 2.5 Pro for guideline generation. 
The LLM analyzes the combined review text to identify common evaluation patterns, important perspectives, and judgment criteria, and then generalizes these findings to generate reviewer guidelines in a single inference step.

The LLM is prompted to generate guidelines by identifying the characteristics of ``good papers'' and ``bad papers.'' 
This process results in guidelines that include evaluation principles such as ``papers with high originality are good'' or ``papers with poor writing quality are bad.'' 
The generated guidelines are formatted as YAML checklists following the structure shown in Listing~\ref{lst:yaml-guideline}, with judgment criteria organized by evaluation category (e.g., clarity, experimental validity) in question format.

For review scoring, we consider two types of guideline formats: rubric-based and non-rubric-based. 
The rubric-based format specifies explicit scoring rules with defined point allocations, whereas the non-rubric-based format provides general evaluation guidance without a fixed scoring framework. 
This approach aims to extract generalized evaluation criteria from large-scale review data and to capture subtle patterns that may not be reflected in manually crafted guidelines.

\section{Prompt of No-guideline}
\label{sec:no-guideline}
For the non-guideline baseline, we used a minimal prompt that did not provide any reviewer guidelines.  
The LLM is simply asked to read the paper and assign an overall rating on a 1–10 scale, along with a brief justification.  
The full prompt is shown in \autoref{lst:non-guideline}.

\begin{lstlisting}[caption={Prompt used in the Non-guideline condition}, label={lst:non-guideline}]
You are an expert paper reviewer. Please read the following paper content and rate its quality.

# Paper
{content}

# Instructions
Provide:
- A numerical rating (1-10) for overall quality.
- A short justification for your score.

# Format
Output the integer overall rating as the following format after your justification:
"Overall Rating: (Your Rating)"
\end{lstlisting}

\section{Prompt of Guideline-based Automated Peer Review}
\label{sec:prompt}
To ensure consistency across different reviewer guidelines, we designed a unified prompt template for automated peer review, as shown in \autoref{lst:peer-review-prompt}.  
Each experiment replaces the placeholder \texttt{\{reviewer\_guideline\}} with the corresponding conference or reviewer-derived guideline (i.e., ICLR, ARR, NeurIPS, and Good/Middle/Bad reviewer style).

\begin{lstlisting}[caption={Prompt template for automated peer review using reviewer guidelines}, label={lst:peer-review-prompt}]
You are an expert peer reviewer. Please read the following paper content and evaluate its quality based on the following reviewer guideline.

# Paper
{content}

# Reviewer Guideline
{reviewer_guideline}

# Final Output
Provide:
- A numerical overall rating (1-10):
	- 10 = Strong Accept
	- 8 = Accept
	- 6 = Weak Accept
	- 5 = Borderline
	- 3 = Weak Reject
	- 1 = Strong Reject
- A justification paragraph that summarizes the key strengths and weaknesses underlying your rating.

# Format
Output the integer overall rating as the following format after your justification:
"Overall Rating: (Your Rating)"
\end{lstlisting}

\paragraph{Official Conference Guidelines}
\label{sec:guideline-sources}
For the official conference guidelines, we used the publicly available reviewer guidelines from each conference’s official website without modification.  
For more details, please refer to the respective conference pages (ICLR\footnote{\url{https://iclr.cc/Conferences/2024/ReviewerGuide}}, 
NeurIPS\footnote{\url{https://neurips.cc/Conferences/2024/ReviewerGuidelines}}, 
and ARR\footnote{\url{https://aclrollingreview.org/reviewerguidelines}}).
We also employed ChatGPT-rewritten versions of these guidelines, reformatted into an instruction style suitable for LLM-based reviewing 
(\autoref{lst:arr-chatgpt}, \autoref{lst:neurips-chatgpt}, and \autoref{lst:iclr-chatgpt}).

\begin{lstlisting}[caption={ARR Reviewer Guideline (ChatGPT-refined version)}, label={lst:arr-chatgpt}]
## Summary
Give a brief summary of the paper in your own words. Highlight the key contributions and main ideas.

## Strengths
List the main strengths of the paper, such as novelty, clarity, strong empirical results, theoretical insights, or broad relevance.

## Weaknesses
Point out the limitations, flaws, or missing components (e.g., weak justification, incomplete experiments, lack of clarity, missing baselines, etc.).

## Assessment by Review Dimension
Please evaluate the paper based on the following criteria from the following guidelines:

- **Soundness/Correctness**: Are the claims supported by the methodology and evidence? Are proofs or derivations correct (if applicable)?
- **Originality/Novelty**: Does the paper offer novel methods, insights, or results? Is the contribution incremental or substantial?
- **Meaningful Comparison**: Does the paper clearly position itself with respect to prior work? Are comparisons (empirical or conceptual) adequate?
- **Clarity**: Is the paper well-written and easy to follow? Are key terms defined and claims clearly stated?
- **Impact/Potential**: Does the paper have the potential to impact future research or practice in its area?
- **Reproducibility**: Are datasets, code, and hyperparameters provided or clearly described? Could others replicate the results?
- **Ethics**: Are there any ethical considerations raised by the paper (e.g., misuse potential, data issues, fairness)? Are these discussed?

You may optionally assign per-dimension scores if helpful, but they are not required.

## Suggestions for Improvement
Provide constructive feedback for the authors to help improve the work, regardless of your score.
---

Please be professional, specific, and constructive. Your goal is to help both the authors and the review community understand the strengths, limitations, and suitability of the paper.
\end{lstlisting}

\begin{lstlisting}[caption={NeurIPS Reviewer Guideline (ChatGPT-refined version)}, label={lst:neurips-chatgpt}]
Your evaluation should be structured into several components.

### Summary
Briefly summarize the paper and its main contributions in your own words. This summary should be a neutral reflection that the authors would agree with, not a critique or a copy of the abstract.

---

### Strengths and Weaknesses
Provide a detailed assessment of the paper's strengths and weaknesses. This section should cover the primary reasons for your recommendation to accept or reject the paper. Evaluate the following dimensions:

* **Quality**: Assess the technical soundness of the submission. Are the claims well-supported by theoretical analysis or experimental results? Are the methods appropriate? Does the work seem complete, and do the authors honestly evaluate both its strengths and weaknesses?
* **Clarity**: Is the paper well-written and organized? Is there enough information for an expert to reproduce the results? Provide constructive suggestions if clarity can be improved.
* **Significance**: How impactful are the results for the machine learning community? Is the work likely to be used or built upon by others? Does it advance the field by addressing a difficult task, providing new insights, or introducing a unique approach?
* **Originality**: Does the paper offer new insights or a deeper understanding of existing methods? How does it differ from previous work, and are relevant citations included? Originality can come from a novel method, a novel combination of techniques, or new insights from evaluating existing methods.

---

### Numerical Ratings for Core Criteria
Based on your assessment in the "Strengths and Weaknesses" section, provide a numerical rating for each of the following categories on a scale from 1 to 4.

* **Quality**:
    * 4: excellent
    * 3: good
    * 2: fair
    * 1: poor
* **Clarity**:
    * 4: excellent
    * 3: good
    * 2: fair
    * 1: poor
* **Significance**:
    * 4: excellent
    * 3: good
    * 2: fair
    * 1: poor
* **Originality**:
    * 4: excellent
    * 3: good
    * 2: fair
    * 1: poor

---

### Questions for the Authors
List 3-5 key, actionable questions or suggestions for the authors. These should focus on points where a response could clarify a confusion, address a limitation, or potentially change your evaluation. Clearly state the criteria under which your score might change.

---

### Limitations and Societal Impact
Assess whether the authors have adequately addressed the limitations of their work and its potential negative societal impact. If they have, a simple "yes" is sufficient. If not, provide constructive feedback for improvement. Remember to reward authors for being transparent about limitations.

\end{lstlisting}

\begin{lstlisting}[caption={ICLR Reviewer Guideline (ChatGPT-refined version)}, label={lst:iclr-chatgpt}]
## Justification and Detailed Review
Write a clear and concise review, addressing each of the following dimensions from the following review guidelines:

- **Strengths and Weaknesses**: Identify the main strengths and weaknesses of the paper.
- **Significance**: Evaluate whether the paper makes a meaningful contribution to the field. Does it advance understanding, methods, or applications in a way that would be interesting to the review community?
- **Originality**: Assess whether the paper presents new ideas, methods, or perspectives.
- **Technical Quality**: Are the methods technically sound and well-justified? Are claims supported by theory or experiments?
- **Clarity**: Is the paper clearly written and well-structured? Can a non-expert in the subfield follow it?
- **Empirical Evaluation** (if applicable): Are experiments well-designed? Are comparisons fair and comprehensive? Are results convincing?
- **Reproducibility**: Does the paper provide enough information (including code/data, if applicable) for others to reproduce the results?
- **Ethics**: Are there any ethical concerns (e.g., societal harm, bias, data misuse)? If yes, how are they handled?

## Confidential Comments to Reviewers (Optional)
(Optional section) Note anything relevant for area chairs, such as borderline decisions, meta-considerations, or conflicts of interest.

---

Be objective, constructive, and concise. Focus on helping the authors and the review community understand the paper's strengths and limitations based on the review standards.
\end{lstlisting}

\paragraph{Reviewer-imitating Guidelines}
\label{sec:reviewer-imitating-guidelines}
We additionally generated reviewer-imitating guidelines from ICLR~2024 reviews to capture common evaluation patterns and characteristic reasoning observed in human reviewers  
(\autoref{lst:good-nonrubric}, \autoref{lst:middle-nonrubric}, and \autoref{lst:bad-nonrubric}). In non-rubric experiments, these guidelines were applied using the standard prompt template shown in \autoref{lst:peer-review-prompt}.
For rubric-based experiments, we used the same reviewer-imitating guidelines (i.e., identical guideline for Good, Middle, and Bad) but replaced the prompt template with a scoring-oriented version (\autoref{lst:rubric-template}) that specifies explicit point-allocation and normalization procedures.

\begin{lstlisting}[caption={Good Reviewer-Imitating Guideline (Non-rubric)}, label={lst:good-nonrubric}]
Provide the following based on the Peer Review Behavior Checklist:

peer_review_behavior_checklist:
  good_paper:
    contribution_and_impact:
      - question: "Does the paper tackle a significant and timely problem?"
        description: "Reviewers positively note papers that address important and relevant research topics, seeing it as a sign of potential impact."
      - question: "Is the core idea clever, insightful, or genuinely novel?"
        description: "A key strength noted is a non-trivial insight or a unique approach that distinguishes the work from incremental improvements."
    clarity_and_presentation:
      - question: "Is the paper exceptionally well-written and easy to follow?"
        description: "Reviewers frequently praise clear writing, logical structure, and understandable explanations as major strengths, even in papers they have concerns about."
      - question: "Are the motivation and contributions communicated unambiguously?"
        description: "A paper is viewed favorably when its purpose and specific contributions are stated clearly from the outset, leaving no room for interpretation."
    technical_and_empirical_soundness:
      - question: "Does the method achieve a significant performance gain over strong, relevant baselines?"
        description: "Convincing, large improvements over well-established and widely-used methods are a primary driver for a positive evaluation."
      - question: "Is the approach technically sound and well-grounded?"
        description: "Reviewers value methods that are built on solid theoretical foundations or have clear, logical justifications for their design choices."
      - question: "Are the experiments thorough, using standard benchmarks and insightful ablations?"
        description: "A comprehensive evaluation across multiple standard datasets, including ablation studies that validate specific design choices, is a hallmark of a strong submission."
      - question: "Are the experimental comparisons conducted fairly?"
        description: "This group of reviewers values direct, fair comparisons where baselines are properly tuned and experimental settings are consistent."
      - question: "Has the author provided source code to support reproducibility?"
        description: "The inclusion of source code is explicitly mentioned as a strength, as it supports the paper's claims and aids reproducibility."
      - question: "Could this work become a standard method or open new research avenues?"
        description: "Reviewers are supportive when they see potential for the work to be widely adopted or to inspire future research."
  bad_paper:
    novelty_and_scholarship:
      - question: "Is the contribution incremental or a re-implementation of existing ideas?"
        description: "This is a frequent and critical flaw. Reviewers are highly knowledgeable and will reject papers they deem to be minor variations of prior work, often citing the specific papers."
      - question: "Does the paper fail to cite or adequately discuss key related work?"
        description: "Missing citations for highly relevant papers is a major red flag, often interpreted as a lack of thoroughness or an attempt to hide a lack of novelty."
    empirical_evaluation:
      - question: "Are the experiments conducted on weak, outdated, or insufficient benchmarks?"
        description: "Evaluation on a very limited set of datasets or non-standard benchmarks is a common reason for rejection, as it fails to prove generalizability."
      - question: "Are the baselines weak, poorly tuned, or missing SOTA competitors?"
        description: "Comparisons are considered unconvincing if the paper omits stronger contemporary methods or appears to have intentionally weakened the performance of its baselines."
      - question: "Do the results actually support the main claims made in the paper?"
        description: "Reviewers critically check if the conclusions drawn in the abstract and introduction are truly backed up by the data presented in the tables and figures."
    clarity_and_soundness:
      - question: "Is the paper poorly written, confusing, or hard to follow?"
        description: "A lack of clarity in writing, notation, or structure is a frequent reason for rejection, as it prevents the reviewer from understanding or verifying the contribution."
      - question: "Are there critical flaws in the methodology, assumptions, or logic?"
        description: "Fundamental errors in the proposed method, unjustified assumptions, or logical gaps in the analysis are considered fatal flaws."
      - question: "Is there enough detail to reproduce the work?"
        description: "Reviewers will reject a paper if it lacks the necessary implementation details, hyperparameters, or architectural information required for another researcher to reproduce the results."
      - question: "Is the problem motivation weak or the contribution's significance unclear?"
        description: "If the paper fails to convincingly argue why the problem is important or why the proposed solution is meaningful, it is often rejected."
      - question: "Are the reported metrics cherry-picked, inappropriate, or presented without context?"
        description: "Reviewers are skeptical of results where the evaluation metrics seem chosen to favor the proposed method or where improvements are marginal without clear statistical significance."
\end{lstlisting}

\begin{lstlisting}[caption={Middle Reviewer-Imitating Guideline (Non-rubric)}, label={lst:middle-nonrubric}]
Provide the following based on the Peer Review Behavior Checklist:

peer_review_behavior_checklist:
  good_paper:
    contribution_and_impact:
      - question: "Is the core idea presented as innovative or a novel application in its domain?"
        description: "Reviewers consistently highlight novelty, being the 'first' to tackle a problem, or proposing a unique approach as a primary reason for acceptance."
      - question: "Is the research problem itself framed as crucial, important, or meaningful?"
        description: "Papers that address a significant and recognized challenge in the field are viewed more favorably, as this gives the work inherent value."
    clarity_and_presentation:
      - question: "Is the paper praised for being well-written, clear, and easy to follow?"
        description: "High writing quality, clear organization, and comprehensible explanations are frequently cited as key strengths, suggesting reviewers value accessibility."
      - question: "Does the paper include an ablation study that clearly justifies each of its components?"
        description: "Reviewers for good papers often look for and praise ablation studies, as they demonstrate a thorough understanding and validation of the proposed architecture."
    evidence_and_validation:
      - question: "Are the experiments described as extensive, thorough, or comprehensive?"
        description: "Strong papers are expected to provide robust empirical validation across multiple datasets, settings, or modalities to solidify their claims."
      - question: "Do the results show a significant and clear improvement over relevant, state-of-the-art baselines?"
        description: "It's not enough to conduct experiments; the results must demonstrate a convincing performance gain against strong competitors to be considered impactful."
      - question: "Does the paper provide theoretical proofs or sound mathematical derivations to support its claims?"
        description: "The inclusion of theoretical grounding is seen as a major strength that adds a layer of rigor and credibility to the empirical results."
      - question: "Are sufficient implementation details or supplementary materials provided to aid understanding and replication?"
        description: "Positive reviews often note the inclusion of clear implementation details or code, which enhances the paper's integrity and perceived value."
    feedback_style:
      - question: "Are the identified weaknesses framed as requests for clarification or minor additions?"
        description: "For papers they intend to accept, reviewers tend to frame weaknesses as constructive questions or suggestions for future work, rather than as critical flaws."
      - question: "Does the reviewer's summary accurately capture the paper's core contributions and strengths?"
        description: "A positive and accurate summary at the beginning of the review often signals that the reviewer has understood and appreciated the paper's main message."
  bad_paper:
    contribution_and_impact:
      - question: "Is the contribution criticized for being incremental, lacking novelty, or a straightforward application of existing work?"
        description: "A perceived lack of novelty is one of the most common and critical reasons for rejection in this review set."
      - question: "Are the paper's central claims or motivation questioned as being unsupported, grandiose, or unconvincing?"
        description: "Reviewers reject papers where the motivation is unclear or the claims are not sufficiently backed by the evidence provided."
    evidence_and_validation:
      - question: "Are the experiments flagged as insufficient, weak, or lacking necessary comparisons?"
        description: "This is a primary reason for rejection. Common issues include not comparing against relevant baselines, using too few datasets, or omitting ablation studies."
      - question: "Are the baseline comparisons criticized as being unfair, weak, or misrepresentative of prior work?"
        description: "Reviewers are critical of evaluations that appear to use outdated or poorly tuned baselines to inflate the perceived performance of the proposed method."
      - question: "Are the reported performance gains described as marginal, insignificant, or within the margin of error?"
        description: "Even when experiments are present, if the improvements are not substantial, the paper is often deemed not significant enough for publication."
    clarity_and_presentation:
      - question: "Is the paper flagged as being hard to follow, confusing, or poorly written?"
        description: "Severe issues with clarity, notation, or organization are critical flaws that prevent reviewers from properly evaluating the work, often leading to rejection."
      - question: "Does the reviewer point out a lack of essential details needed for understanding or replication?"
        description: "The omission of crucial methodological or experimental details is treated as a major flaw that undermines the paper's credibility."
    methodological_soundness:
      - question: "Are there fundamental flaws identified in the proposed method's design, assumptions, or logic?"
        description: "Reviewers will recommend rejection if they find deep-seated problems in the core methodology that invalidate the approach."
      - question: "Is the method criticized for being overly complex or a 'laundry list' of disconnected components?"
        description: "Papers proposing convoluted models without clear justification for their complexity are viewed negatively."
      - question: "Is the method's robustness questioned, for instance, due to high sensitivity to hyperparameters?"
        description: "A lack of analysis on hyperparameter sensitivity or robustness is a common weakness cited in negative reviews."periments, theory, or statistically significant results (e.g., overlapping confidence intervals)."
\end{lstlisting}

\begin{lstlisting}[caption={Bad Reviewer-Imitating Guideline (Non-rubric)}, label={lst:bad-nonrubric}]
Provide the following based on the Peer Review Behavior Checklist:

peer_review_behavior_checklist:
  good_paper:
    contribution_and_impact:
      - question: "Does the paper achieve state-of-the-art or highly competitive results?"
        description: "Reviewers in this group frequently highlight strong empirical performance against established benchmarks as a primary strength and a key reason for acceptance."
      - question: "Is the problem being addressed important and relevant to the community?"
        description: "Papers are viewed more favorably when they tackle a problem that is clearly articulated as significant and timely for the field."
      - question: "Is the core idea novel, interesting, and well-motivated?"
        description: "Novelty and clear motivation are consistently praised. Reviewers look for a contribution that is not just an incremental tweak but offers a new perspective or useful framework."
    presentation_and_clarity:
      - question: "Is the paper well-written, clearly organized, and easy to follow?"
        description: "High value is placed on clarity. Reviews for good papers often explicitly mention that the writing is good and the method is easy to understand."
      - question: "Does the paper use effective visualizations to explain its core ideas?"
        description: "The use of clear, informative visualizations is noted as a significant strength that enhances the reader's understanding of the proposed concepts."
    evaluation_and_rigor:
      - question: "Are the experiments comprehensive and the evaluations thorough?"
        description: "Accepted papers are often praised for having sound, comprehensive experiments with solid ablation studies that justify their claims and design choices."
      - question: "Is the study complete and the analysis insightful?"
        description: "Reviewers appreciate papers that seem 'complete' in their investigation, providing a full picture rather than a preliminary exploration."
      - question: "Is the work reproducible, for instance, by open-sourcing code and data?"
        description: "While not a universal requirement, providing code is explicitly mentioned as a positive factor that increases the paper's value and reproducibility."
      - question: "Is the proposed solution reasonable and well-justified for the problem?"
        description: "A logical and promising connection between the problem statement and the proposed solution is a key characteristic of papers that receive positive feedback."
      - question: "Does the paper clearly outperform relevant baselines?"
        description: "Demonstrating a clear and significant improvement over existing, well-chosen baselines is a very common and persuasive strength."
  bad_paper:
    contribution_and_novelty:
      - question: "Is the contribution just a combination of existing methods with limited novelty?"
        description: "This is the most common and critical flaw identified. Papers are frequently criticized for lacking originality or simply stitching together known components."
      - question: "Is the contribution too simple, minor, or incremental to be significant?"
        description: "Reviewers often reject papers where the core idea, while perhaps functional, is deemed too simplistic or its contribution too marginal for a top-tier venue."
      - question: "Is the motivation for the work unclear or is the problem framing unconvincing?"
        description: "A frequent weakness is the failure to clearly explain why the problem is important or why the proposed approach is necessary."
    experiments_and_evidence:
      - question: "Are the experiments weak, insufficient, or conducted on overly simple datasets?"
        description: "Papers are consistently penalized for weak empirical validation, such as using toy datasets, not running enough experiments, or lacking sufficient detail like random seeds."
      - question: "Does the paper fail to compare against necessary SOTA or strong baseline methods?"
        description: "An inadequate comparison to the state of the art or relevant baselines is a major red flag and a common reason for rejection."
      - question: "Are key design choices left unsubstantiated by ablation studies?"
        description: "Reviewers expect authors to justify components of their method through ablations; the absence of these studies is a frequently cited weakness."
      - question: "Are the claims exaggerated or not sufficiently supported by the evidence provided?"
        description: "A mismatch between the claims made in the text and the actual results presented in the experiments is a critical flaw."
    clarity_and_soundness:
      - question: "Is the paper poorly written, hard to follow, or filled with jargon and typos?"
        description: "Poor presentation, including unclear language, logical gaps, and typos, is a significant factor in negative reviews."
      - question: "Does the method lack theoretical justification, proofs, or guarantees?"
        description: "For papers proposing new algorithms, the absence of theoretical analysis or justification is frequently pointed out as a major weakness."
      - question: "Does the paper show a lack of awareness of relevant prior work?"
        description: "Missing citations to important and recent related work is seen as a sign of poor scholarship and is a common point of criticism."
\end{lstlisting}

\begin{lstlisting}[caption={Rubric-based Template for Reviewer-Imitating Guidelines}, label={lst:rubric-template}]

You are an expert peer reviewer. Please read the following paper content and evaluate its quality based on the following instructions.

# Paper
{content}

# Reviewer Guideline
{reviewer_imitating_guideline}

# Scoring Procedure:
- For each of the 10 questions in the `good_paper` section:
  - If the answer is Yes, assign +1 point, else 0.
- For each of the 10 questions in the `bad_paper` section:
  - If the answer is Yes, assign -1 point, else 0.

Compute the temporary score by summing all the above responses.
- This results in a value between -10 and +10.

Normalize the temporary score to a section rating in the range 1 to 10 using the following formula:
normalized_score = round(((temporary_score + 10) / 20) * 9 + 1)

This maps:
- -10 -> 1  
- 0 -> 5 
- +10 -> 10  

# Section Rating

Provide:
- The temporary score (-10 to +10)
- The normalized section rating (1 to 10)
- A short explanation of the score based on your answers to the checklist

# Final Output

Provide:
- A numerical integer overall rating that is same as the normalized section rating (1-10):
    - 10 = Strong Accept  
    - 8 = Accept  
    - 6 = Weak Accept  
    - 5 = Borderline  
    - 3 = Weak Reject  
    - 1 = Strong Reject  

- A justification paragraph summarizing the key strengths and weaknesses of the paper

Finally, write the final score using the following exact format (on its own line):
"Overall Rating: (Your Rating)"
\end{lstlisting}

\section{Qualitative Textual Analysis of LLM-Generated Reviews}
\label{sec:analysis}

While RMSE effectively measures numerical alignment, it serves as a narrow metric that does not fully capture the quality, helpfulness, or logical depth of the textual justifications. To provide a more multidimensional view of how guidelines influence the granularity of automated reviews, we present a qualitative textual analysis. \autoref{tab:analysis} summarizes the differences in review text generated for a single sample paper (ID: \textit{0bjIoHD45G}) by the Qwen3-30B-A3B model, contrasting the outputs produced under official NeurIPS guidelines (shown in \autoref{lst:neurips_output}) versus the non-rubric ``Bad Reviewer'' imitating guidelines (shown in \autoref{lst:bad_output}. The raw generated texts (with internal model reasoning traces removed for brevity) are provided below the table.

\begin{lstlisting}[caption={Official NeurIPS Guideline Output for Paper 0bjIoHD45G}, label={lst:neurips_output}]
The paper presents a novel approach to bridge the performance gap between deep learning (DL) and tree-based methods on tabular data by addressing implicit categorical features and leveraging Fourier embeddings. While the work is well-structured and experimentally rigorous, several critical issues in the checklist raise concerns about reproducibility, transparency, and completeness.

**Strengths:**
- **Clear Contributions:** The paper clearly states its goals of identifying implicit categorical features and using Fourier embeddings to improve DL performance on tabular data. The experiments demonstrate significant improvements over baselines like XGBoost and MLP/RESNET.
- **Detailed Experimental Setup:** The methodology includes hyperparameter random search, multiple tasks (classification/regression), and a comprehensive benchmark with 68 datasets. The paper describes preprocessing steps, model architectures (MLP, RESNET), and evaluation metrics (accuracy, R^2, MAE).
- **Empirical Validity:** The results show that the proposed methods (ResNet+F|C, MLP+F|C) outperform baselines on several tasks, with "spiking" performance gains in specific cases, suggesting the effectiveness of the approach.

**Weaknesses:**
- **Lack of Code/Data Availability:** The paper does not explicitly mention releasing code, data, or detailed instructions for reproducibility (Question 5). While the appendix may contain supplementary material, the main text does not clarify this, which is critical for verifying results.
- **Missing Statistical Significance Metrics:** The paper reports average performance but lacks error bars, confidence intervals, or statistical tests to validate the significance of results (Question 7). This weakens the reliability of claims about performance improvements.
- **No Limitations Section:** The paper does not include a dedicated "Limitations" section, despite acknowledging that the work is preliminary. This omission makes it harder to assess the scope and robustness of the findings (Question 2).
- **Incomplete Experimental Details:** While the paper describes the general setup, specific hyperparameters (e.g., learning rates, batch sizes) and computational resources (e.g., GPU types, memory) are not detailed in the main text, limiting reproducibility (Questions 6 and 8).
- **Ethical and Societal Impact Gaps:** The paper does not address potential negative societal impacts, which is required by the checklist (Question 10). While tabular data tasks are less likely to have direct harms, this omission is still a concern.

**Overall Rating: 6 (Weak Accept)** The paper makes a valuable contribution to the field of tabular data learning, with strong empirical results and a clear methodology. However, critical gaps in reproducibility, statistical rigor, and ethical considerations prevent it from meeting the highest standards of transparency and completeness required for a stronger acceptance. Addressing these issues would significantly strengthen the paper's impact and credibility.  

Overall Rating: 6
\end{lstlisting}

\begin{table*}[t!]
\small
\centering
\begin{tabular}{p{0.2\linewidth} p{0.35\linewidth} p{0.35\linewidth}}
\toprule
\textbf{Feature} & \textbf{Official NeurIPS Guideline Output} & \textbf{``Bad Reviewer'' Guideline Output} \\
\midrule
\textbf{Tone \& Structure} & Highly structured. Explicitly divides the review into bulleted \textbf{Strengths} and \textbf{Weaknesses}. The tone is objective and grounded in specific submission requirements. & Unstructured. Presents a single, monolithic block of text. The tone is generally positive but relies on vague, generalized critiques. \\
\addlinespace
\textbf{Actionability of Critique} & Very high. Points out concrete, fixable gaps: lack of a dedicated Limitations section, missing code/data links, absence of error bars/statistical tests, and missing ethical/societal impact statements. & Low. Relies on generalized tropes typical of poor human reviews, such as citing ``limited theoretical justification'' and vague ``formatting issues'' without specific examples. \\
\addlinespace
\textbf{Score Calibration} & \textbf{6 (Weak Accept)}. The score is well-calibrated to the text; the model acknowledges strong empirical results but penalizes the paper appropriately for critical missing elements required by top-tier venues. & \textbf{8 (Accept)}. The score is poorly calibrated to the text. The model assigns a strong Accept rating despite explicitly noting a lack of code, limited theoretical justification, and inconsistent results across tasks. \\
\bottomrule
\end{tabular}
\caption{Qualitative Comparison of Generated Justifications for Paper 0bjIoHD45G}
\label{tab:analysis}
\end{table*}

\begin{lstlisting}[caption={``Bad Reviewer'' Guideline Output for Paper 0bjIoHD45G}, label={lst:bad_output}]
The paper presents a well-structured approach to addressing the challenges of applying deep learning to tabular data by introducing methods for identifying implicitly categorical features and leveraging Learned Fourier Features (LFF) to mitigate smoothness bias.

The experiments are comprehensive, with thorough comparisons against strong baselines like XGBoost and detailed ablation studies. The core ideas-statistical tests for implicit categorical features and LFF embeddings-are novel and well-motivated, addressing a significant gap in the literature.

The paper's contributions are impactful, particularly in demonstrating that deep learning models can achieve competitive performance on tabular data with proper preprocessing.

However, the lack of open-sourced code and data, limited theoretical justification for LFF in this context, and some formatting issues slightly detract from its overall quality. The empirical results are strong but not universally consistent across all tasks, and the paper could benefit from more detailed ablation studies and a broader comparison with state-of-the-art deep learning methods.

Overall Rating: 8
\end{lstlisting}

As demonstrated in \autoref{tab:analysis}, guideline design drastically alters the utility and logical depth of the LLM's output. The official NeurIPS guidelines prompted the model to generate a structured, balanced evaluation that constructively weighed both the merits and flaws of the methodology. Because the official guidelines mandate checking for specific scientific rigors (e.g., reproducibility, statistical significance, ethics), the model provided highly actionable feedback. Conversely, the ``Bad Reviewer'' guideline, which was distilled from low-quality human reviews prioritizing behavioral tropes, resulted in an unconstructive, unstructured critique. Furthermore, the ``Bad Reviewer'' prompt caused a disconnect between the textual critique and the final rating, highlighting how vague guidelines can lead to poorly calibrated automated scoring. This qualitative difference underscores our quantitative finding that explicit, conference-refined criteria are essential for effective LLM-based peer review.

\begin{figure*}[h]
    \centering
    \includegraphics[width=\linewidth]{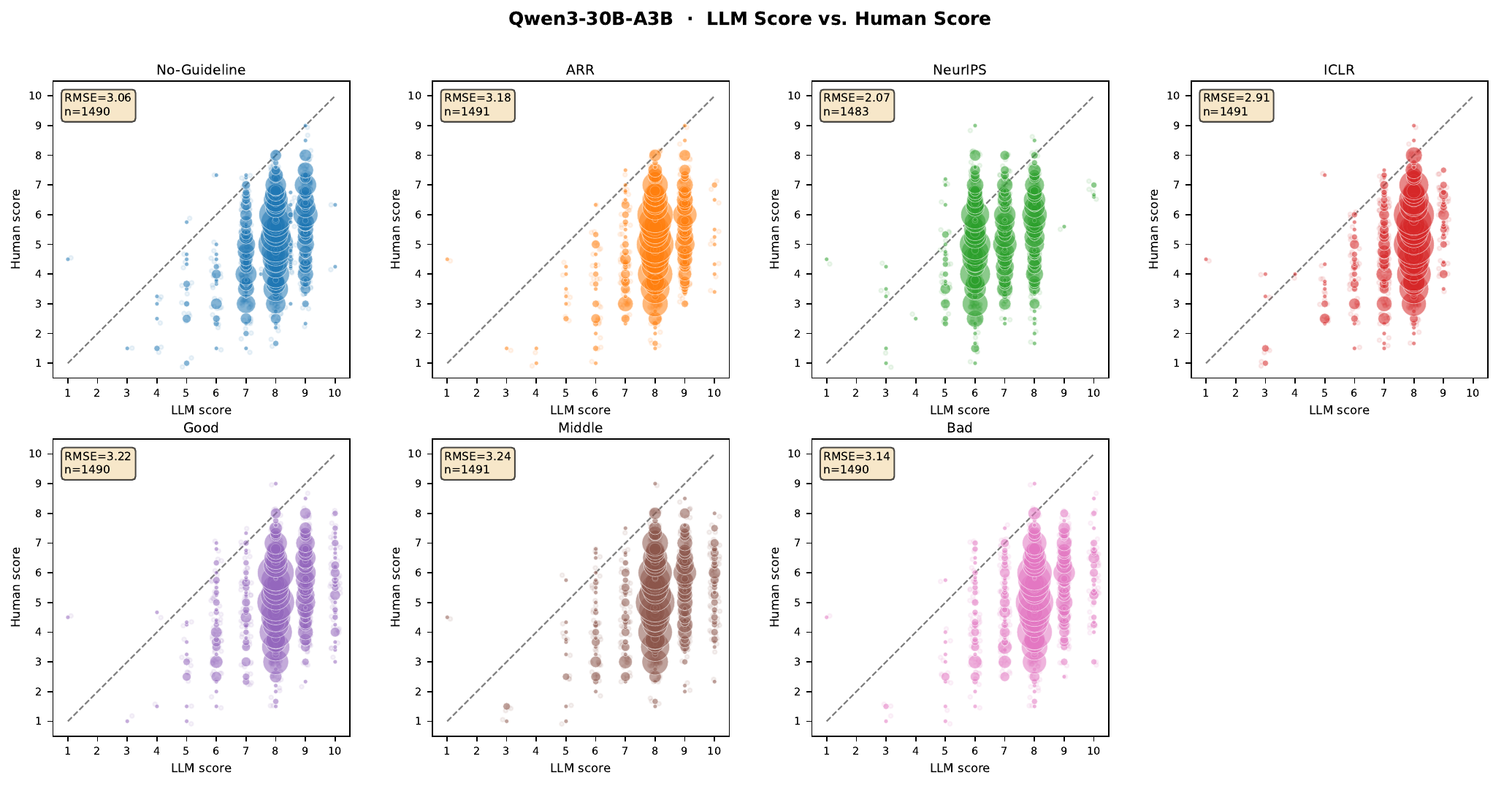} 
    \caption{Scatter plots comparing LLM scores to human scores for the Qwen3-30B-A3B model across various guideline conditions. The ``No-Guideline'' condition exhibits severe score centralization, which is mitigated when structured guidelines (e.g., NeurIPS) are applied.}
    \label{fig:scatter_plots}
\end{figure*}

\begin{figure*}[h]
    \centering
    \includegraphics[width=0.8\linewidth]{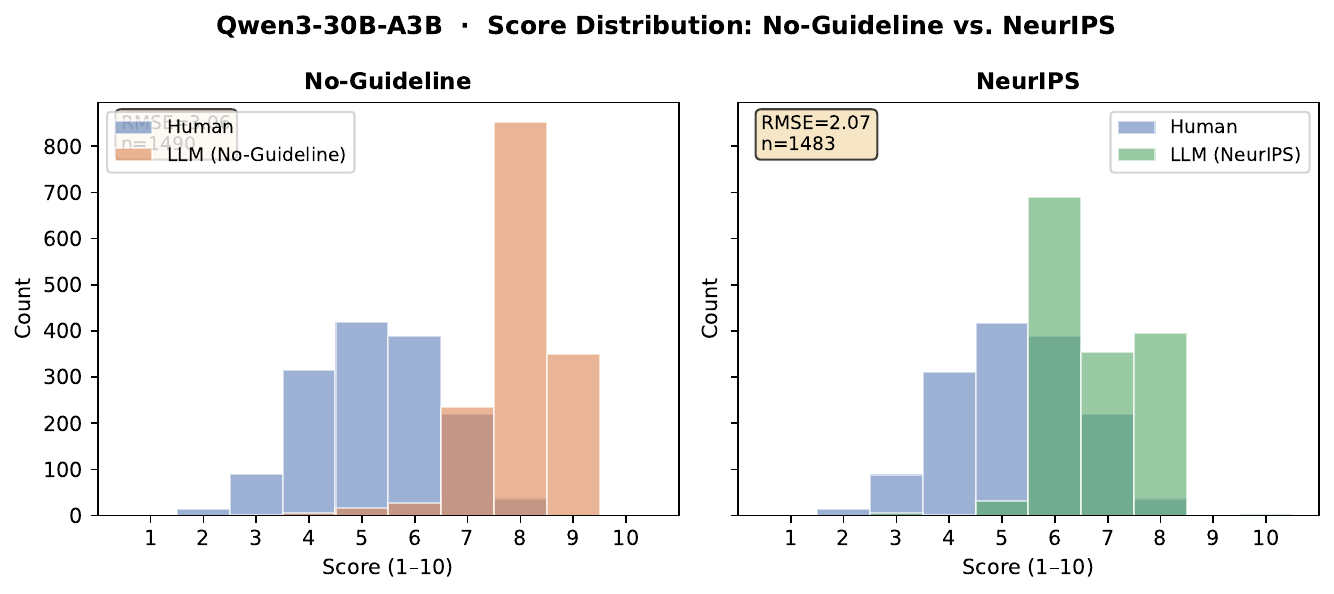} 
    \caption{Histograms comparing the score distributions of the No-Guideline baseline against the NeurIPS guideline for the Qwen3-30B-A3B model. The baseline shows an unnatural spike at the mean, whereas the NeurIPS guideline yields a distribution that better approximates human judgments.}
    \label{fig:histograms}
\end{figure*}

\section{Visualization of Score Distributions}
\label{sec:score_distributions}

To address potential concerns regarding whether LLMs take shortcuts by defaulting to mean scores, we visualize the score distributions produced by our models. \autoref{fig:scatter_plots} presents scatter plots comparing LLM predictions to human averages across different guideline conditions. \autoref{fig:histograms} shows the corresponding score frequencies for the No-Guideline baseline and the official NeurIPS guidelines.

As shown in the visualizations, the ``No-Guideline'' baseline suffers from severe score centralization. The model frequently collapses its predictions to the dataset's mean score (acting as a safe, default guess), which is vividly evident in the flat horizontal clustering in the scatter plot and the massive unnatural spike in the histogram. However, when the model is provided with explicit, well-structured criteria (such as the official NeurIPS guidelines), the generated scores spread much more naturally across the 1-10 scale. The scatter plots under the official guidelines show a much stronger alignment with the diagonal of human judgments. This confirms that high-quality reviewer guidelines effectively prevent the model from taking shortcuts, forcing it to engage in granular, paper-specific evaluation.

\end{document}